%% file: ms.tex
\documentclass[10pt,twocolumn,letterpaper]{article}

\usepackage{iccv}
\usepackage{times}
\usepackage{epsfig}
\usepackage{graphicx}
\usepackage{amsmath}
\usepackage{amssymb}
\usepackage{booktabs}
\usepackage{subcaption}
\usepackage{multirow}
\usepackage{comment}

\graphicspath{{./figures/}}
\input{./macrosjvg}

\usepackage[pagebackref=true,breaklinks=true,letterpaper=true,colorlinks,bookmarks=false]{hyperref}

\iccvfinalcopy 

\def\iccvPaperID{1673} 

\ificcvfinal\fi
\begin{document}

\title{Active Decision Boundary Annotation with Deep Generative Models }


\author{
Miriam Huijser\\
Aiir Innovations\\
Amsterdam, The Netherlands \\
{\tt\small \url{https://aiir.nl/} }
\and
Jan C. van Gemert\\
Delft University of Technology\\
Delft, The Netherlands\\
{\tt\small \url{http://jvgemert.github.io/} }
}

\maketitle
\pagenumbering{gobble}

\input{abstract}
\input{intro}

\input{related}

\input{method}

\input{experiments}

\input{conclusion}

{\small
\bibliographystyle{ieee}
\bibliography{./egbib}
}

\input{1673-supp}

\end{document}

%% file: macrosjvg.tex
\usepackage{xcolor}
\newcommand{\todo}[1]{[{\bf \color{red} TODO: #1}]}

\newcommand{\Eq}[1]{Eq.~(\ref{eq:#1})}

\newcommand{\fig}[1]{figure~\ref{fig:#1}}
\newcommand{\tab}[1]{Table~\ref{tab:#1}}

\def\argmax{\operatornamewithlimits{\rm arg\,max}}

\usepackage{booktabs}

%% file: abstract.tex
\begin{abstract}
This paper is on \emph{active learning} where the goal is to reduce the data annotation burden by interacting with a (human) oracle during training. Standard active learning methods ask the oracle to annotate data samples. Instead, we take a profoundly different approach: we ask for annotations of the decision boundary. We achieve this using a deep generative model to create novel instances along a 1d line. A point on the decision boundary is revealed where the instances change class. Experimentally we show on three data sets that our method can be plugged into other active learning schemes, that human oracles can effectively annotate points on the decision boundary, that our method is robust to annotation noise, and that decision boundary annotations improve over annotating data samples.
\end{abstract}

%% file: intro.tex
\section{Introduction}

If data is king, then annotation labels are its crown jewels. Big image data sets are relatively easy to obtain; it's the ground truth labels that are expensive~\cite{dengCVPR09imagenet, linECCV14mscoco}. With the huge success of deep learning methods critically depending on large annotated datasets, there is a strong demand for reducing the annotation effort~\cite{jacobsen2016structured, kingma2014semi, mettes2016spot, owensECCV16ambientSupervision, pathakCVPR16contextEncoder}. 

In \textit{active learning}~\cite{settles2010active} the goal is to train a good predictor while minimizing the human annotation effort for large unlabeled data sets. During training, the model can interact with a human oracle who provides ground truth annotations on demand. The challenge is to have the model automatically select a small set of the most informative annotations so that prediction performance is maximized.

\begin{figure}
\centering
\includegraphics[width=\linewidth]{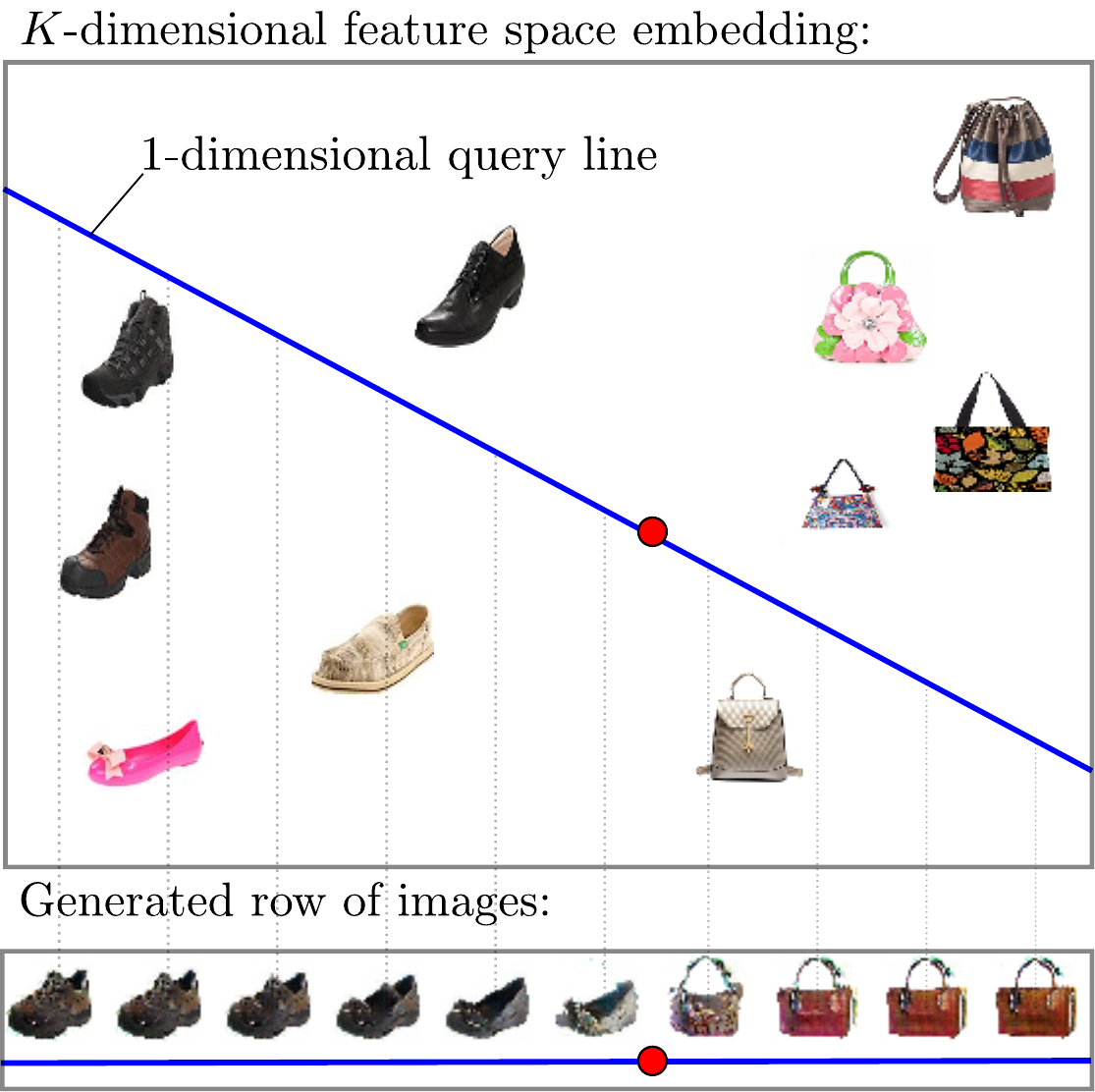}
\caption{Active decision boundary annotation using a deep generative model. In a K-dimensional feature space (top), a 1-dimensional  query line (blue) is converted to a row of images (bottom) by generating visual samples along the line. The oracle annotates where the generated samples change classes (red point). This point lies close to the decision boundary and we use that point to improve classification.}
\label{fig:visualAbstract}
\end{figure}
  
In this paper we exploit the power of deep generative models for active learning. Active learning methods~\cite{lewis1994sequential, settles2010active, tong2001support} typically ask the oracle to label an existing data sample. Instead, we take a radically different approach: we ask the oracle to directly annotate the decision boundary itself. To achieve this, we first use all unlabeled images to learn a K-dimensional embedding. In this K-dimensional embedding we select a 1-dimensional query line and employ a deep generative model to generate visual samples along this line. We visualize the 1-dimensional line as an ordered row of images and simply ask the oracle to visually annotate the point where the generated visual samples changes classes. Since the generated images are ordered, the oracle does not need to examine and annotate each and every image, merely identifying the change point is enough. The point between two samples of different classes is a point that lies close to the decision boundary and we use that point to improve the classification model. In \fig{visualAbstract} we show an illustration of our approach.

We make the following contributions. First, we use a deep generative model to present a 1-dimensional query line to the oracle. Second, we directly annotate the decision boundary instead of labeling data samples. Third, we learn a decision boundary from both labeled instances and boundary annotations. Fourth, we evaluate if the generative model is good enough to construct query lines that a human oracle can annotate, how much noise the decision boundary annotations allow, and how our decision boundary annotation version of active learning compares to traditional active learning where data samples are labeled.

%% file: related.tex
\section{Related work}

Active learning~\cite{settles2010active} is an iterative framework and starts with the initialization of a prediction model either by training on a small set of labeled samples or by random initialization. Subsequently, a \emph{query strategy} is used to interactively query a oracle which can be another model or a human annotator. The annotated query is then used to retrain the model and starts the next iteration. Active learning in computer vision includes work on selecting the most influential images~\cite{freytag2014selectingInfluential}, refraining from labeling unclear visuals~\cite{kading2015activeUnnameable}, zero-shot transfer learning~\cite{GavvesICCV15activeTransfer}, multi-label active learning~\cite{wang2016multiLabelActive}. Similar to these methods, our paper uses active learning in the visual domain to minimize the number of iterations while maximizing prediction accuracy. 

There are several settings of active learning. A \emph{pool-based} setting~\cite{lewis1994sequential,tong2001support} assumes the availability of a large set of unlabeled instances.  Instead, a \emph{stream-based} setting ~\cite{atlas1989training,cohnML94stream} is favorable for  online learning where a query is selectively sampled from an incoming data stream. Alternatively, in a \emph{query synthesis} setting~\cite{angluin1988queries, baum1992query, zhu2017generativeAdversarialActiveLearning} an input data distribution is used to generate queries for the oracle in the input space. Recently, \cite{alabdulmohsin2015efficient} and \cite{chen2016dimension} proposed methods for efficiently learning halfspaces, i.e. linear classifiers, using synthesized instances.  Instead of using an existing setting, our paper proposes a new active learning setting: \emph{active boundary annotation}. Other active learners use sample instances to query the oracle. Instead, we generate a row of instances and query the point where the instances change class label. We do not query an annotation of a sample, we query an annotation of the decision boundary.


Query strategies in active learning are the informativeness measures used to select or generate new queries. Much work has been done on this topic~\cite{settles2010active}, here we describe a few prominent strategies. \emph{Uncertainty sampling}~\cite{lewis1994sequential} is a query strategy that selects the unlabeled sample of which the current model is least certain. This could be obtained by sampling closest to the decision boundary~\cite{campbell2000queryLearnLargeMargin, tong2001support} or based on entropy~\cite{lewis1994heterogeneousEntropy}. The \emph{Uncertainty-dense sampling} method~\cite{settles2008analysis,zhu2010activeDense}  aims to correct for the problems associated with uncertainty sampling by selecting samples that are not only uncertain but that also lie in dense areas of the data distribution and are thus representative of the data. In \emph{Batch} methods~\cite{guo2008discriminative, shen2005active, xu2007incorporating}, not a single sample is queried at each iteration, but a set of samples. In \emph{Query-by-committee}~\cite{seung1992query} multiple models are used as a committee and queries the samples where the disagreement is highest. Our active boundary annotation method can plug-in any sampling method and thus does not depend on a particular query strategy. We will experimentally demonstrate that our boundary annotation method can readily be applied to various query strategies.

In our active learning approach we make use of deep generative models. Probabilistic generative models such as variational autoencoders~\cite{kingma2014semi,kingma2013auto}, learn an inference model to map images to a latent space and a decoder to map from latent space back again to image space. Unfortunately the generated images are sometimes not very sharp. Non-probabilistic models such as the generative adversarial nets (GAN)~\cite{goodfellow2014generative} produce higher-quality images than variational autoencoders~\cite{NIPS2015_5773}, but can only map from latent space to image space. These models randomly sample latent vectors from a predefined distribution and then learn a mapping from these latent vectors to images; there is no mapping that can embed real images in the latent space. To perform classification on real images in the embedding we need an inference step to map images to the latent space. Fortunately, recent generative adversarial models can produce high-quality images and provide efficient inference \cite{donahue2016adversarial,dumoulin2016adversarially}. In this paper we use such a GAN model.


%% file: method.tex
\section{Active decision boundary annotation}

\begin{figure}
\centering
\includegraphics[width=\linewidth]{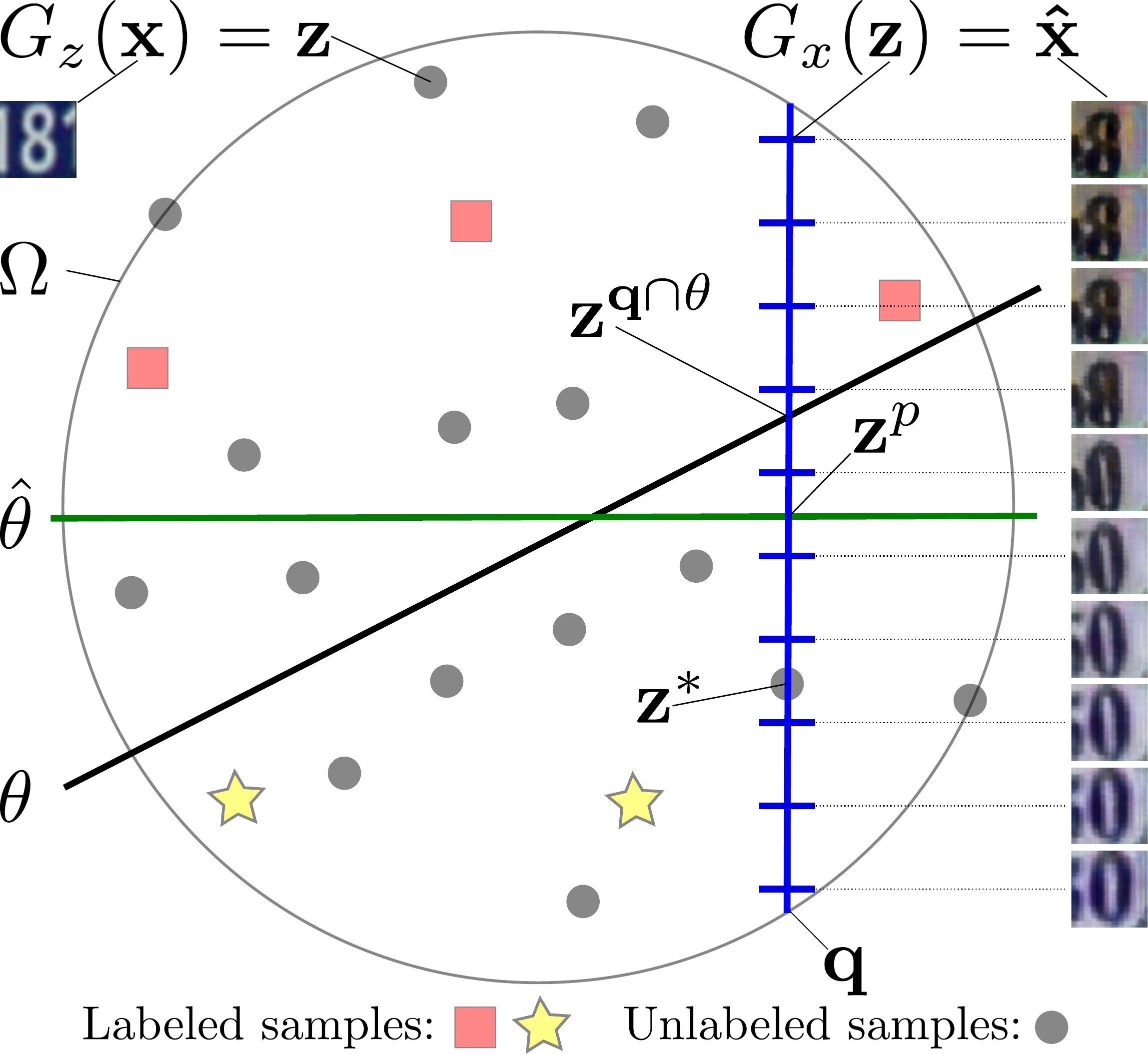}
\caption{Overview of our method. Data labels $y$ are the red squares and yellow stars; unlabeled data is a gray circle.  A deep generative model is used to map an image sample $\mathbf{x}$ to its corresponding latent variable $G_z(\mathbf{x}) = \mathbf{z}$. Vice versa, an image is generated from the latent variable $\mathbf{z}$ by $G_x(\mathbf{z})=\mathbf{\hat x}$. The hypersphere $\Omega$ bounds the latent space. The true decision boundary (black line) is $\theta$ and the current estimate of the decision boundary (green line) is $\hat \theta$. The query line $\mathbf{q}$ (blue) goes through the query sample  $\mathbf{z}^*$ and is perpendicular to the current estimate of the decision boundary $\hat \theta$, intersecting it at point $\mathbf{z}^p$. The query line $\mathbf{q}$ is uniformly sampled (blue bars) and bounded by $\Omega$, which gives a row of generated images as illustrated at the right. Note how a `0' morphes to an `8' after it passes the decision boundary $\theta$. The latent boundary annotation point is given by $\mathbf{z}^{ \mathbf{q} \cap \theta }$. }
\label{fig:method}
\end{figure}

We have $N$ data samples $\mathbf{x} \in \mathbb{R}^D$, each sample is paired with a class label $(\mathbf{X,Y}) = \{ (\mathbf{x}_1, y_1), \ldots,(\mathbf{x}_N, y_N) \}$ where for clarity we focus on the binary case $y \in \{-1,1\}$ and a linear classification model.
As often done in active learning we assume an initial set $\mathcal{A}$ which contains a handful of annotated images.
Each data sample $\mathbf{x}_i$ has a corresponding latent variable $\mathbf{z}_i \in \mathbb{R}^K$. For clarity we omit the index $i$ whenever it is clear that we refer to a single data point. The tightest hypersphere that contains all latent variables $\mathbf{z}_i$ is denoted by $\Omega$. Every iteration in active learning estimates a decision boundary $\hat \theta$ where the goal is to best approximate the real decision boundary $\theta$ while minimizing the number of iterations.

In \fig{method} we show an overview of our method.  Each image $\mathbf{x}$ is embedded in a manifold as a latent variable $G_z(\mathbf{x}) = \mathbf{z}$. In this embedding we use a standard active learning query strategy to select the most informative query sample $\mathbf{z}^*$. We then construct a query line $\mathbf{q}$ in such a way that it intersects the query sample $\mathbf{z}^*$ and is perpendicular to the current estimate of the decision boundary $\hat \theta$ at point $\mathbf{z}^p$. We uniformly sample  latent points $\mathbf{z}$ along the 1-dimensional query line $\mathbf{q}$ and for each point on the line generate an estimate of the corresponding image $G_x(\mathbf{z})=\mathbf{\hat x}$. On this generated row of images we ask the oracle to provide the point where the images change class, this is where the decision boundary intersects the query line $\mathbf{q} \cap \theta$ and the latent variable $\mathbf{z}^{ \mathbf{q} \cap \theta }$ is a decision boundary annotation. Using the boundary annotation we can assign a label to the query sample $\mathbf{z}^*$ which we add to the set  of annotated samples $\mathcal{A}$. All annotated decision boundary points are stored in the set $\mathcal{B}$. The estimate of the decision boundary $\hat \theta$ is found by optimizing a joint classification/regression loss. The classification loss is computed on the labeled samples $\mathcal{A}$ while at the same time the regression aims to fit the decision boundary through the annotations in $\mathcal{B}$. 

\textbf{Deep generative embedding.} We make use of GANs (Generative Adversarial Nets)~\cite{goodfellow2014generative} to obtain a high-quality embedding. In GANs, a generative model $G$ can create realistic-looking images from a latent random variable $G(\mathbf{z})=\mathbf{\hat x}$. The generative model  is trained by making use of a strong discriminator $D$ that tries to separate synthetic generated images from real images. Once the discriminator cannot tell synthetic images from real images, the generator is well trained. The generator and discriminator are simultaneously trained by playing a two-player minimax game, for details see~\cite{goodfellow2014generative}.

\textbf{Deep generative inference.} Because we perform classification in the embedding we need an encoder to map the image samples to the latent space. Thus, in addition to a mapping from latent space to images (decoding) as in a standard GAN, we also need an inference model to map images to latent space (encoding)~\cite{donahue2016adversarial, dumoulin2016adversarially}. This is done by optimizing two joint distributions over $(\mathbf{x,z})$: one for the encoder $q(\mathbf{x},\mathbf{z}) = q(\mathbf{x})q(\mathbf{z}|\mathbf{x})$ and one for the decoder $p(\mathbf{x},\mathbf{z})=p(\mathbf{z})p(\mathbf{x}|\mathbf{z})$. Since both marginals are known, we can sample from them. The encoder marginal $q(\mathbf{x})$ is the empirical data distribution and the decoder marginal is defined as $p(\mathbf{z}) = \mathcal{N}(\mathbf{0},\mathbf{I})$. The encoder and the decoder are trained together to fool the discriminator. This is achieved by playing an adversarial game to try and match the encoder and decoder joint distributions where the adversarial game is played between the encoder and decoder on the one side and the discriminator on the other side, for details see~\cite{dumoulin2016adversarially}.

\textbf{Query strategy.} For compatibility with standard active learning methods we allow plugging in any query strategy that selects an informative query sample $\mathbf{z}^*$. Such a plug-in is possible by ensuring that the query sample $\mathbf{z}^*$ is part of the query line $\mathbf{q}$. An example of a popular query sample strategy is \emph{uncertainty sampling} that selects the sample whose prediction is the least confident: $\mathbf{z}^* = \argmax_\mathbf{z} 1 - P_{\hat \theta}(\hat y | \mathbf{z})$, where $\hat y$ is the class label with the highest posterior probability under the current prediction model $\hat \theta$. This can be interpreted as the sample where the model is least certain about. We experimentally validate our method's compatibility with common query strategies.  

\textbf{Constructing the query line.} The query line $\mathbf{q}$ determines which images will be shown to the oracle. To make decision boundary annotation possible these images should undergo a class-change. A reasonable strategy is then to make sure the query line intersects the current estimate of the decision boundary $\hat \theta$. A crisp class change will make annotating the point of change easier. Thus, we increase the likelihood of a crisp class-change by ensuring that $\mathbf{q}$ is perpendicular to $\hat \theta$. Since the query sample $\mathbf{z}^*$ lies on the query line and $\mathbf{q} \perp \hat \theta$ this is enough information to construct the query line $\mathbf{q}$. Let the current estimation of the linear decision boundary $\hat \theta$ be a hyperplane which is parametrized by a vector $\mathbf{\hat w}$ perpendicular to the plane and offset with a bias $\hat b$. Then the query line is given by $\mathbf{q}(t) = \mathbf{z}^p + (\mathbf{z}^* - \mathbf{z}^p)t$, where $\mathbf{z}^p$ is the projection of the query point $\mathbf{z}^*$ on the current decision boundary $\hat \theta$ and is defined as $\mathbf{z}^p = \mathbf{z}^* - \frac{(\mathbf{\hat w}^\intercal\mathbf{z}^* + \hat b)}{\mathbf{\hat w}^\intercal\mathbf{\hat w}}  \mathbf{\hat w}$, for derivation details see Appendix~\ref{app:appendix_projection}.

\textbf{Constructing the image row.} We have to make a choice about which image samples along the query line $\mathbf{q}$ to show to the oracle. We first restrict the size of $\mathbf{q}$ to lie within the tightest hypersphere $\Omega$ that captures all latent data samples. The collection of points $\mathcal{H}$ that lie within and on the surface of the hypersphere $\Omega$ is defined as $\mathcal{H} = \{ \mathbf{z} \in \mathbb{R}^K: || \mathbf{z} - \bar{ \mathbf{z}}  || \leq r  \} $ where $r = \max_{i=0}^{N} || \mathbf{z_i} - \bar{ \mathbf{z}} ||$ and $\bar{ \mathbf{z}}$ is the average vector over all latent samples $\mathbf{z}$.  As a second step we uniformly sample $s$ latent samples from $\mathbf{q} \cap \mathcal{H}$ which are decoded to images with the deep generative model $G(\mathbf{z})=\mathbf{\hat x}$.

\textbf{Annotating the decision boundary.} Annotating the decision boundary can be done by a human oracle or by a given model. A model is often used as a convenience to do large scale experiments where human experiments are too time-consuming. For example, large scale experiments for active learning methods that require query sample annotation are typically performed by replacing the human annotation by the ground truth annotation. This assumes that the human will annotate identically to the ground truth, which may be violated close to the decision boundary. To do large scale experiments for our decision boundary annotation method we also use a model based on the ground truth, like commonly done for query sample annotation. We train an oracle-classifier on ground truth labels and use the that model as a ground truth decision boundary where the intersection  $\mathbf{z}^{ \mathbf{q} \cap \theta } =  \mathbf{q}  \cap \theta$
 can be computed. For both oracle types --the oracle-classifier and the human oracle-- we store the decision boundary annotations in $\mathcal{B}$ and we also ask the annotators for the label $y$ of the query point $\mathbf{z}^*$, for which we store the pair $(\mathbf{z}^*,y)$ in $\mathcal{A}$.  We experimentally evaluate human and model-based annotations.

\textbf{Model optimization using boundary annotations.} At every iteration of active learning we update $\hat \theta$.  We use  a classification loss on the labeled samples in $\mathcal{A}$ while at the same time we optimize a regression loss to fit the decision boundary through the annotations in $\mathcal{B}$. In this paper we restrict ourselves to the linear case and parametrize  $\hat \theta$ with a linear model $\hat{ \mathbf{w}}^\intercal \mathbf{z} + \hat b = 0$. For the classification loss we use a standard SVM hinge loss over the labeled samples $(\mathbf{z},y)$ in $\mathcal{A}$ as
\begin{equation}
\mathcal{L}_\text{class}  =  \frac{1}{|\mathcal{A}|} \sum_{(\mathbf{z},y) \in \mathcal{A}} \max\left( 0, 1-y (\mathbf{\hat w}^\intercal \mathbf{z} + \hat b) \right).
\end{equation}
For the regression, we use a simple squared loss to fit the model $\hat w + \hat b$ to the annotations in $\mathcal{B}$
\begin{equation}
\mathcal{L}_\text{regress}  =  \frac{1}{|\mathcal{B}|} 
 \sum_{\mathbf{z} \in \mathcal{B}} 
 \left(\mathbf{\hat w}^\intercal \mathbf{z} + \hat b \right)^2.
\end{equation}
The final loss $\mathcal{L}$ jointly weights the classification and regression losses equally and simply becomes
\begin{equation}
\mathcal{L} = \frac{1}{2}\mathcal{L}_\text{class} + 
\frac{1}{2} \mathcal{L}_\text{regress} + \lambda ||\mathbf{\hat w}||^2,
\end{equation}
where the parameter $\lambda$ controls the influence of the regularization term $||\mathbf{\hat w}||^2$ where we use $\lambda=1$ in all experiments. Note that because $\mathcal{L}_\text{class}$ and $\mathcal{L}_\text{regress}$ are both convex losses, the joint loss $\mathcal{L}$ is convex as well.

%% file: experiments.tex
\begin{table*}
\begin{tabular}{l ll ll ll}
 \multicolumn{7}{c}{Experiment 1: Evaluating various query strategies} \\ \toprule
 & \multicolumn{2}{c}{MNIST 0 vs. 8} & \multicolumn{2}{c}{SVHN 0 vs. 8} & \multicolumn{2}{c}{Shoe-Bag} \\
Strategy & Sample & Boundary (ours) &   Sample & Boundary (ours) &   Sample & Boundary (ours)  \\ \cmidrule(lr){1-1} \cmidrule(lr){2-3} \cmidrule(lr){4-5} \cmidrule(lr){6-7}
 Uncertainty  & $144.0 \pm 0.5$ & \textbf{145.8 $\pm$ 0.4} & $118.7 \pm 1.3$ & \textbf{124.3 $\pm$ 1.0} & $143.2 \pm 0.6$ & \textbf{145.4 $\pm$ 0.5} \\
 Uncertainty-dense &   $135.6 \pm 10.5$ & \textbf{142.0 $\pm$ 10.8} & $99.6 \pm 5.8$ & \textbf{116.8 $\pm$ 2.5} & $112.0 \pm 6.6$ & \textbf{135.2 $\pm$ 3.0} \\
5 Cluster centroid &  $141.7 \pm 0.4$ & \textbf{145.0 $\pm$ 0.3} & $98.0 \pm 4.9$ & \textbf{106.3 $\pm$ 1.6} &  $131.0 \pm 1.6 $& \textbf{143.7 $\pm$ 0.3} \\
Random & $142.2 \pm 1.0$ & \textbf{145.1 $\pm$ 0.5} & $116.2 \pm 1.9$ & \textbf{124.7 $\pm$ 1.1} & $140.5 \pm 1.1$ & \textbf{145.0 $\pm$ 0.4}\\
\bottomrule
\end{tabular}
\caption{ AULC results for four active learning query strategies. Results are on MNIST (classifying 0 and 8), SVHN (classifying 0 and 8) and Shoe-Bag after 150 queries, where the maximum possible AULC score is 150. The results are averaged over 15 repetitions. For each row, the significantly best result is shown in bold, where significance is measured with a paired t-test with p $<$ 0.05. SVHN is the most difficult dataset. Uncertainty sampling is generally the best query strategy. Boundary annotation significantly outperforms sample annotations for all datasets for all query strategies.}
\label{tab:queryStrategies}
\end{table*}

\section{Experiments}
We perform active learning experiments on three datasets. MNIST contains 60,000 binary digit images, 50k to train and 10k in the test set. The SVHN dataset~\cite{netzer2011reading} contains challenging digit images from Google streetview, it has 73,257 train and 26,032 test images. The SVHN dataset has 531,131 extra images, which we use to train the embedding. In the third dataset we evaluate our method on more variable images than digits. We create the Shoe-Bag dataset of 40,000 train and 14,000 test images by taking subsets from the Handbags dataset~\cite{zhu2016generative} and the Shoes dataset~\cite{yu2014fine}. 

For every dataset we train a deep generative embedding following~\cite{dumoulin2016adversarially}. For MNIST and SVHN we train an embedding with 100 dimensions, for the more varied Shoe-Bag we train an embedding of 256 dimensions. For the training of the embeddings we set dropout = 0.4 for the layers of the discriminator. All embeddings are trained on the train data, except for the SVHN embedding; which is trained on the larger ``extra'' dataset following~\cite{dumoulin2016adversarially}. We will make our code available~\cite{code}. All learning is done in the embedding and the experiments that do not use a human oracle use an SVM trained on all labels as the oracle.

We evaluate active learning with the Area Under the (accuracy) Learning Curve (AULC) measure~\cite{o2016model,settles2008analysis}.  
The Area under the Learning Curve is computed by integrating over the test accuracy scores of $N$ active learning iterations using the trapezoidal rule:
\begin{equation}
\text{AULC} = \sum_{i=1}^N \frac{1}{2}(\text{acc}_{i-1} + \text{acc}_i),
\end{equation}
where $\text{acc}_0$ is the test accuracy of the initial classifier before the first query.
The AULC score measures the informativeness per annotation and is high for active learning methods that quickly learn high-performing models with few queries, i.e. in few iterations. We also report Mean Average Precision (MAP) results in Appendix~\ref{app:resultsInAP}.

We first evaluate essential properties of our method on  two classes: `0' versus `8'. Later we evaluate on all classes. 


\begin{figure*}
\centering
   \begin{subfigure}[b]{0.96\textwidth}
   \includegraphics[width=1\linewidth,clip,trim= 10mm 2mm 5mm 4mm]{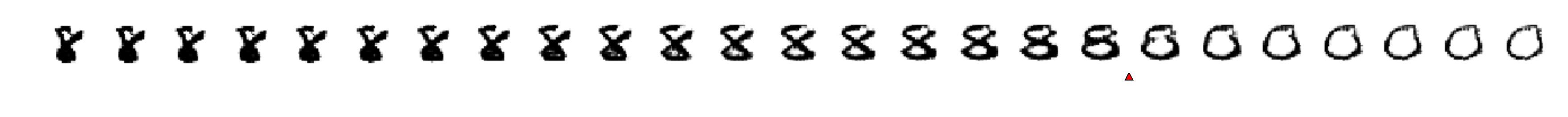}
   \includegraphics[width=1\linewidth,clip,trim= 10mm 2mm 5mm 4mm]{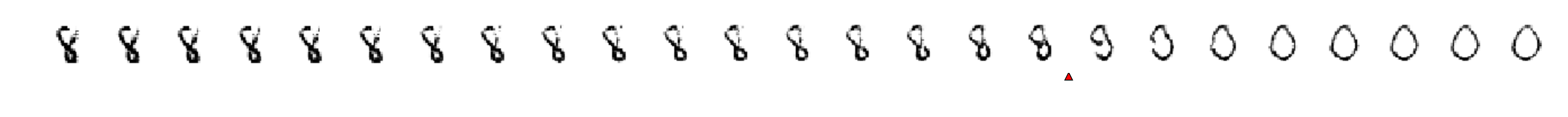}
   \includegraphics[width=1\linewidth,clip,trim= 10mm 2mm 5mm 4mm]{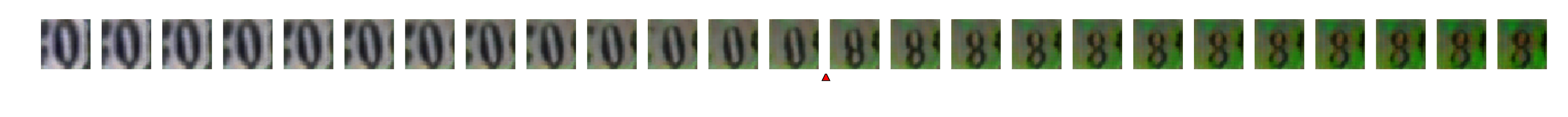}
   \includegraphics[width=1\linewidth,clip,trim= 10mm 2mm 5mm 4mm]{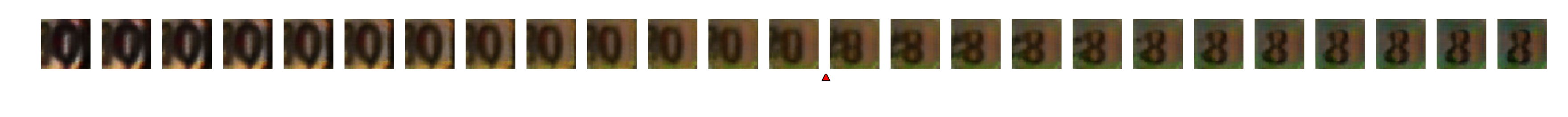}
   \includegraphics[width=1\linewidth,clip,trim= 10mm 2mm 5mm 4mm]{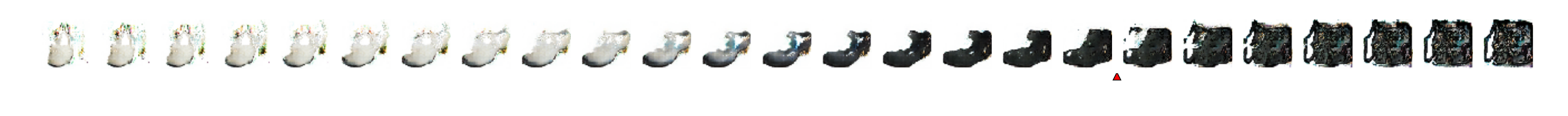}
   \includegraphics[width=1\linewidth,clip,trim= 10mm 2mm 5mm 4mm]{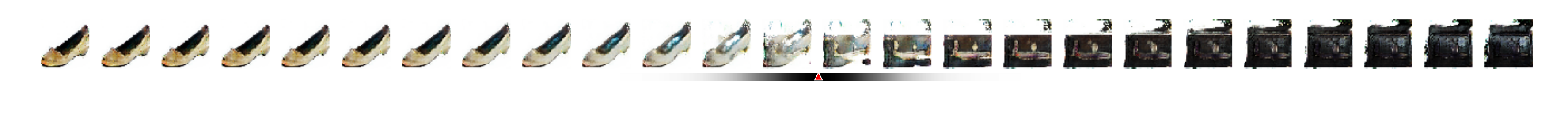}
   \caption{Query lines with high human consistency.}
   \label{fig:goodlines}
\end{subfigure}

\begin{subfigure}[b]{0.96\textwidth}
   \includegraphics[width=1\linewidth,clip,trim= 10mm 2mm 5mm 4mm]{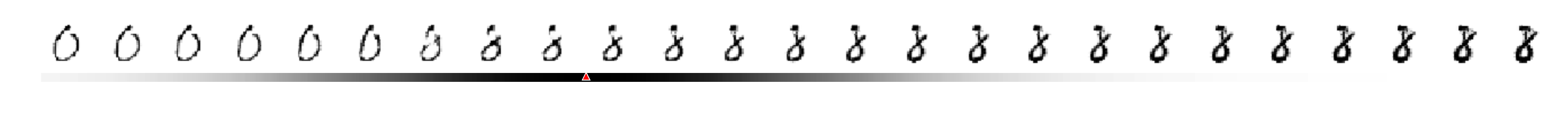}
   \includegraphics[width=1\linewidth,clip,trim= 10mm 2mm 5mm 4mm]{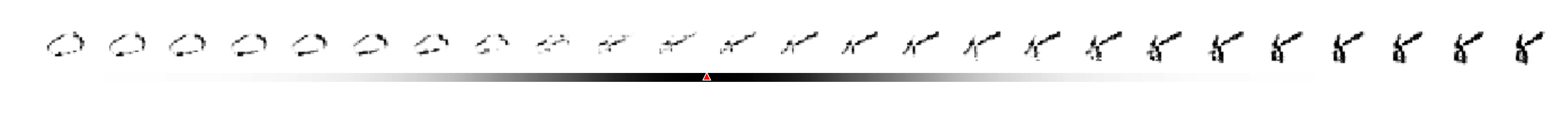}
   \includegraphics[width=1\linewidth,clip,trim= 10mm 2mm 5mm 4mm]{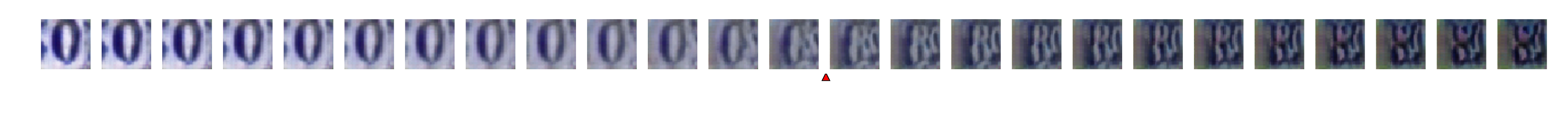}
   \includegraphics[width=1\linewidth,clip,trim= 10mm 2mm 5mm 4mm]{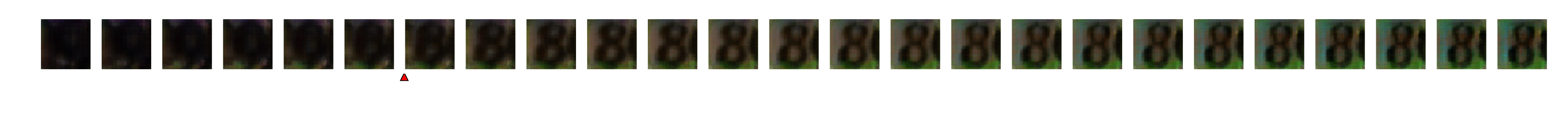}
   \includegraphics[width=1\linewidth,clip,trim= 10mm 2mm 5mm 4mm]{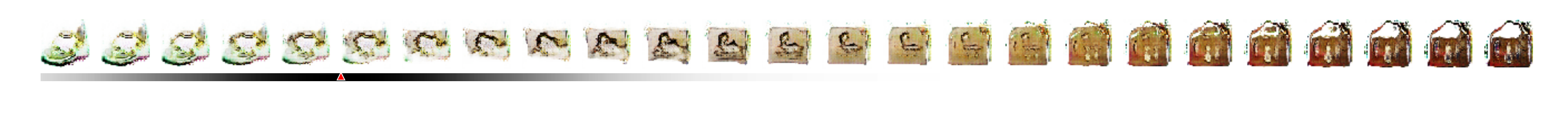}
   \includegraphics[width=1\linewidth,clip,trim= 10mm 3mm 5mm 4mm]{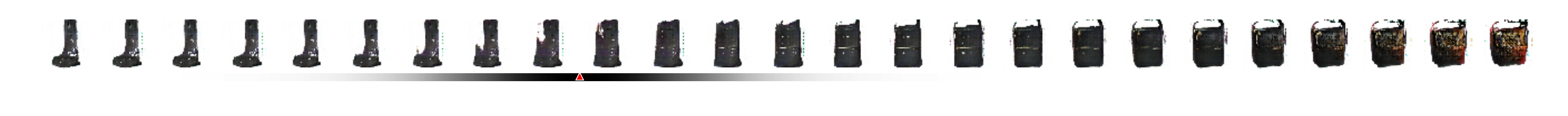}
   \caption{Query lines with low human consistency.}
   \label{fig:badlines}
\end{subfigure}
\caption{ Examples of line queries for MNIST (top two rows) SVHN (middle two rows) and shoe-bag (bottom two rows). Red triangles indicate the mean annotation per line and the gray bar indicates the standard deviation from the mean annotation. (a) Query lines for which the 10 human annotators were most consistent and (b) query lines for which the human annotators were most inconsistent. For visibility these query lines are subsampled in 14 images; the human annotators were presented with more or fewer images depending on the length of the line query. The human annotators are more consistent for query lines with clearer images and a more sudden change of class, such as the third and fourth row from the top. It should be noted, however, that the class-changes on these lines are not as sudden as is visualized here; the human annotators were presented with more images, also seeing the images in between the images presented here.}
\label{fig:lines_human_annotation}
\end{figure*}

\subsection{Exp 1: Evaluating various query strategies}
\label{sec:traditionalAL}

Our method can re-use standard active learning query strategies that select a sample for oracle annotation. We evaluate four sample-based query strategies. \emph{Random sampling} is a common baseline for query strategies~\cite{ramirez2017active}. \emph{Uncertainty sampling} selects the least confident sample point~\cite{lewis1994sequential}. \emph{Uncertainty-dense sampling}~\cite{settles2008analysis} selects samples that are not only uncertain but that also lie in dense areas of the data distribution. The \emph{$K$-cluster centroid}~\cite{shen2005active} uses batches of $K$ samples, where we set $K=5$. We plug in each query sample strategy in our line query construction approach used for decision boundary annotation.  

The results in~\tab{queryStrategies} show that for all three datasets and for all four query strategies our boundary annotations outperform sample annotations. For the uncertainty-dense method the improvement is the largest, which may be due to this method sampling from dense areas of the distribution, and boundary annotations add complementary information. The uncertainty sampling gives best results for both active learning methods and all three datasets. It is also the strategy where our method improves the least, and is thus the most challenging to improve upon. We select uncertainty sampling for the remainder of the experiments.

\subsection{Exp 2: Evaluating generative model quality}
The generative model that we plug into our method should be able to construct recognizable line queries so that human oracles can annotate them. In~\fig{lines_human_annotation} we show some line queries generated for all three datasets by our active learning method with uncertainty sampling. Some query lines are difficult to annotate, as shown in~\fig{lines_human_annotation}(b) and others are of good quality as shown in~\fig{lines_human_annotation}(a). 

We quantitatively evaluate the generation quality per dataset by letting 10 humans each annotate the same 10 line queries. Line queries are subsampled to have a fixed sample resolution of $0.25$, i.e. the distance between each image on the line and thus vary in length depending on their position in the hypersphere $\Omega$.  The human annotators are thus presented with more images for longer query lines and fewer images for shorter query lines. 
For all 10 line queries we evaluate the inter-human annotation consistency. A higher consistency suggests that the query line is well-recognizable and thus that the generative model has a good image generation quality. We instructed the human oracles to annotate the point where they saw a class change; or indicate if they see no change, this happens for 8 out of the 30 lines.

\begin{table}
\begin{tabular}{l c c}
 \multicolumn{3}{c}{Experiment 2: Evaluating inter-human annotations} \\ \toprule
 & lines without change  & samples deviation \\
\cmidrule(lr){2-2} \cmidrule(lr){3-3} 
MNIST 0 vs. 8  & 2  & $4$\\
SVHN 0 vs. 8   & 1  & $1$\\
Shoe-Bag       & 5  & $9$\\
\bottomrule
\end{tabular}
\caption{Annotation consistency results averaged over 10 query line annotations from 10 human oracles. We show the number of lines marked as having no class change and the  average deviation in number of images, rounded up, from the average annotation per line. Human consistency is worse for the non-uniform Shoe-Bag dataset. The more uniform datasets MNIST and SVHN have quite accurate human consistency. }
\label{tab:humanAnnotators}
\end{table}

In~\tab{humanAnnotators} we show the results for the inter-human annotation consistency.  The Shoe-Bag embedding does not seem to be properly trained because the human annotators see no change in half of the query lines. In addition, the variance between the images make the consistency lower. MNIST has a deviation of 4 images and 2 lines were reported with no change. SVHN provides the highest quality query lines - the human annotators agreed collectively on the inadequacy of only one query line and the human annotators are most consistent for this dataset.

\begin{table*}
\begin{tabular}{l ll ll ll} 
\multicolumn{7}{c}{Experiment 3: Evaluating annotation noise} \\ \toprule
 Sampling noise& \multicolumn{2}{c}{MNIST 0 vs. 8} & \multicolumn{2}{c}{SVHN 0 vs. 8} & \multicolumn{2}{c}{Shoe-Bag} \\
(\# images) & Sample & Boundary (ours) &   Sample & Boundary (ours) &   Sample & Boundary (ours)  \\ \cmidrule(lr){1-1} \cmidrule(lr){2-3} \cmidrule(lr){4-5} \cmidrule(lr){6-7}
0 & $144.2 \pm 0.5$ & \textbf{146.0 $\pm$ 0.3} & $119.1 \pm 1.5$ & \textbf{124.0 $\pm$ 0.9} & $143.1 \pm 0.6$ &  \textbf{145.4 $\pm$ 0.5}\\
1 &$144.2 \pm 0.5$ & \textbf{145.9 $\pm$ 0.3} & $119.1 \pm 1.5$& \textbf{123.4 $\pm$ 1.1} & $143.1 \pm 0.6$ & \textbf{145.2 $\pm$ 0.4}\\
2 & $144.2 \pm 0.5$ &\textbf{145.4 $\pm$ 0.5} & $119.1 \pm 1.5$ & \textbf{121.4 $\pm$ 2.1} & $143.1 \pm 0.6$ & \textbf{144.7 $\pm$ 0.9}\\
3 &$144.2 \pm 0.5$ & \textbf{145.0 $\pm$ 0.4} &$119.1 \pm 1.5$& \textbf{121.1 $\pm$ 1.2} & $143.1 \pm 0.6$& \textbf{144.5 $\pm$ 0.7}\\
4 & $144.2 \pm 0.5$ &$144.2 \pm 0.4$ &$119.1 \pm 1.5$& $119.1 \pm 0.9$                   & $143.1 \pm 0.6$ & \textbf{143.9 $\pm$ 0.5}\\
5 & \textbf{144.2 $\pm$ 0.5} & $143.6 \pm 0.4$ & $119.1 \pm 1.5$ & $113.6 \pm 10.7$ & $143.1 \pm 0.6$ & $143.0 \pm 0.7$\\
\bottomrule
\end{tabular}
\caption{AULC results for noisy boundary active learning with uncertainty sampling for MNIST (classifying 0 and 8), SVHN (classifying 0 and 8) and Handbags vs. Shoes after 150 queries (maximum possible score is 150). Each experiment is repeated 15 times. For each row, the significantly best result is shown in bold, where significance is measured with a paired t-test with p $<$ 0.05. Noise has been added to the boundary annotation points; not to the image labels. Results worsen with more added noise, with the turning point of the significant better performance of Boundary around a sampling noise of 4 images for MNIST and SVHN, and 5 images for Shoe-Bag.}
\label{tab:labelnoise}
\end{table*}

\begin{table*}
\centering 
\begin{tabular}{l ll ll ll} 
\multicolumn{7}{c}{Experiment 4: Evaluating a human oracle} \\ \toprule
  & \multicolumn{2}{c}{MNIST 0 vs. 8} & \multicolumn{2}{c}{SVHN 0 vs. 8} & \multicolumn{2}{c}{Shoe-Bag} \\
Annotation & Sample & Boundary (ours) &   Sample & Boundary (ours) &   Sample & Boundary (ours)  \\ \cmidrule(lr){1-1} \cmidrule(lr){2-3} \cmidrule(lr){4-5} \cmidrule(lr){6-7}
Human oracle & $8.5 \pm 0.7$& \textbf{8.8 $\pm$ 0.3} &  $5.7 \pm 0.4$& $5.8 \pm 0.4$  & $8.1 \pm 0.5$ & $8.2 \pm 0.4$ \\
SVM oracle & $8.9 \pm 0.3$& \textbf{9.1$\pm$ 0.3}& $6.3 \pm 0.4$& $6.4 \pm 0.4$& $8.7 \pm 0.3$ & $8.8 \pm 0.4$  \\ 
\bottomrule
\end{tabular}
\caption{AULC results for a human and a SVM oracle for sample-based active learning and our boundary active learning for MNIST (classifying 0 and 8), SVHN (classifying 0 and 8) and Shoe-Bag after 10 queries (maximum possible score is 10). The experiments are repeated 15 times and  significant results per row are shown in bold for p $<$ 0.05.  Results always improve for boundary annotation, but these improvements are not significant for SVHN and Shoe-Bag.  }
\label{tab:humanOracle}
\end{table*}

\subsection{Exp 3: Evaluating annotation noise}

In experiment 2 we show that there is variation in the annotations between human oracles. Here we aim to answer the question if that variation matters. We evaluate the effect of query line annotation noise on the classification performance. We vary the degree of additive line annotation noise with respect to SVM oracle decision boundary annotations on the 1-dimensional line query.  
We vary the standard deviation $\sigma$ of Gaussian noise to $\sigma \in \{1, \ldots, 5 \}$ image samples away from the oracle.

The results in~\tab{labelnoise} show that adding sampling noise of up to about $\sigma=4$ images to the SVM oracle annotations has a slight negative effect on the performance of our boundary annotation method, but it is still significantly better than sample annotation. Comparing these results to the inter-human annotation consistency results in~\tab{humanAnnotators} shows that Shoe-Bag annotation variation is around $9$, and thus the quality of the generator will likely degrade accuracy. For MNIST and SVHN the human consistency is around or below 4 images which is well-handled.

%

\subsection{Exp 4: Evaluating a human oracle}

In this experiment we evaluate classification performance with a human oracle. For all three datasets we have a human oracle annotate the first 10 line queries, selected using uncertainty sampling. We repeat the same experimental setup for sample-based active learning. The results are averaged over 15 repetitions. 

\begin{figure*}
\centering
   \begin{subfigure}{0.33\textwidth}
   \includegraphics[width=1\linewidth]{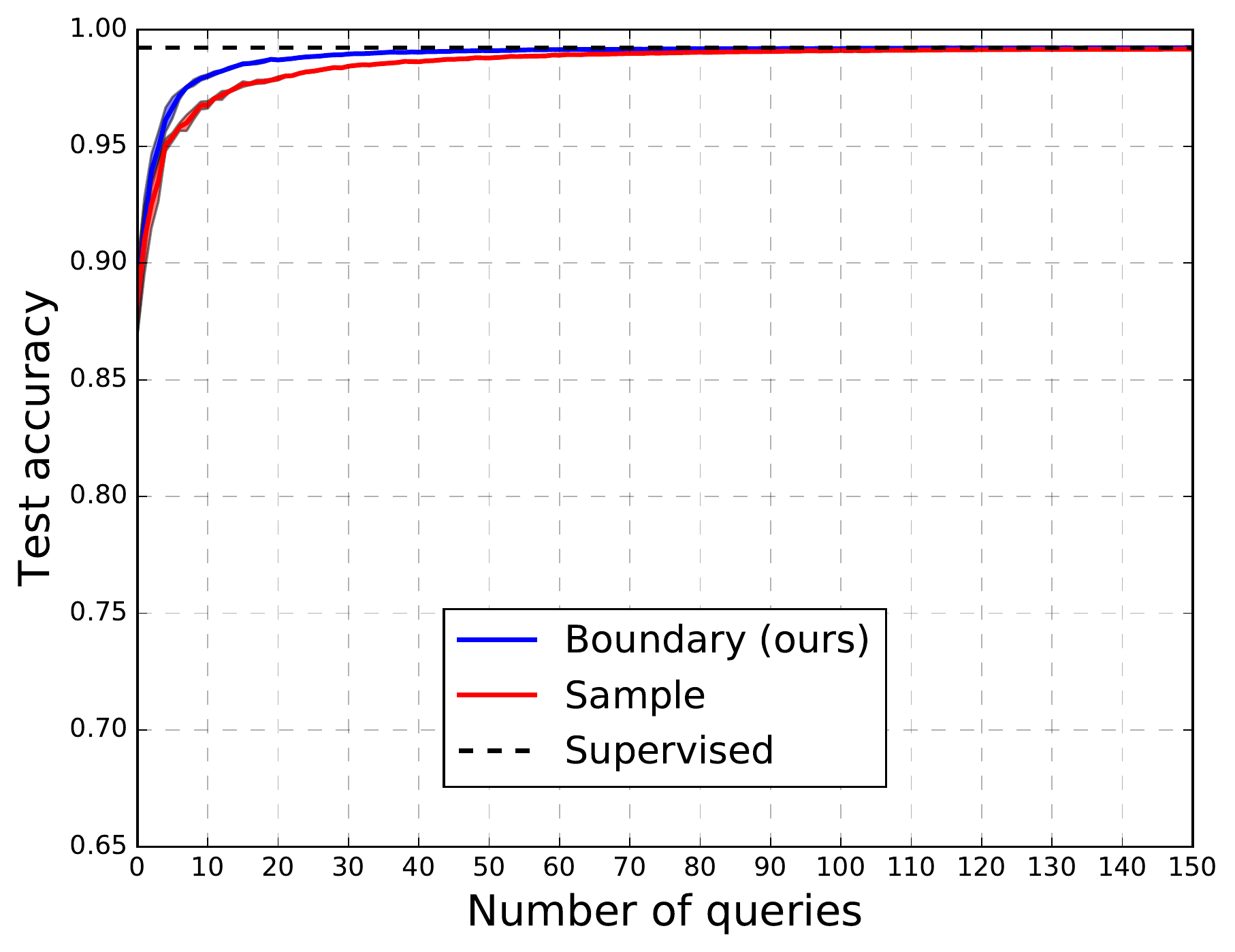}
   \caption{MNIST averaged over all classes.}
   \label{}
\end{subfigure}
\begin{subfigure}{0.33\textwidth}
   \includegraphics[width=1\linewidth]{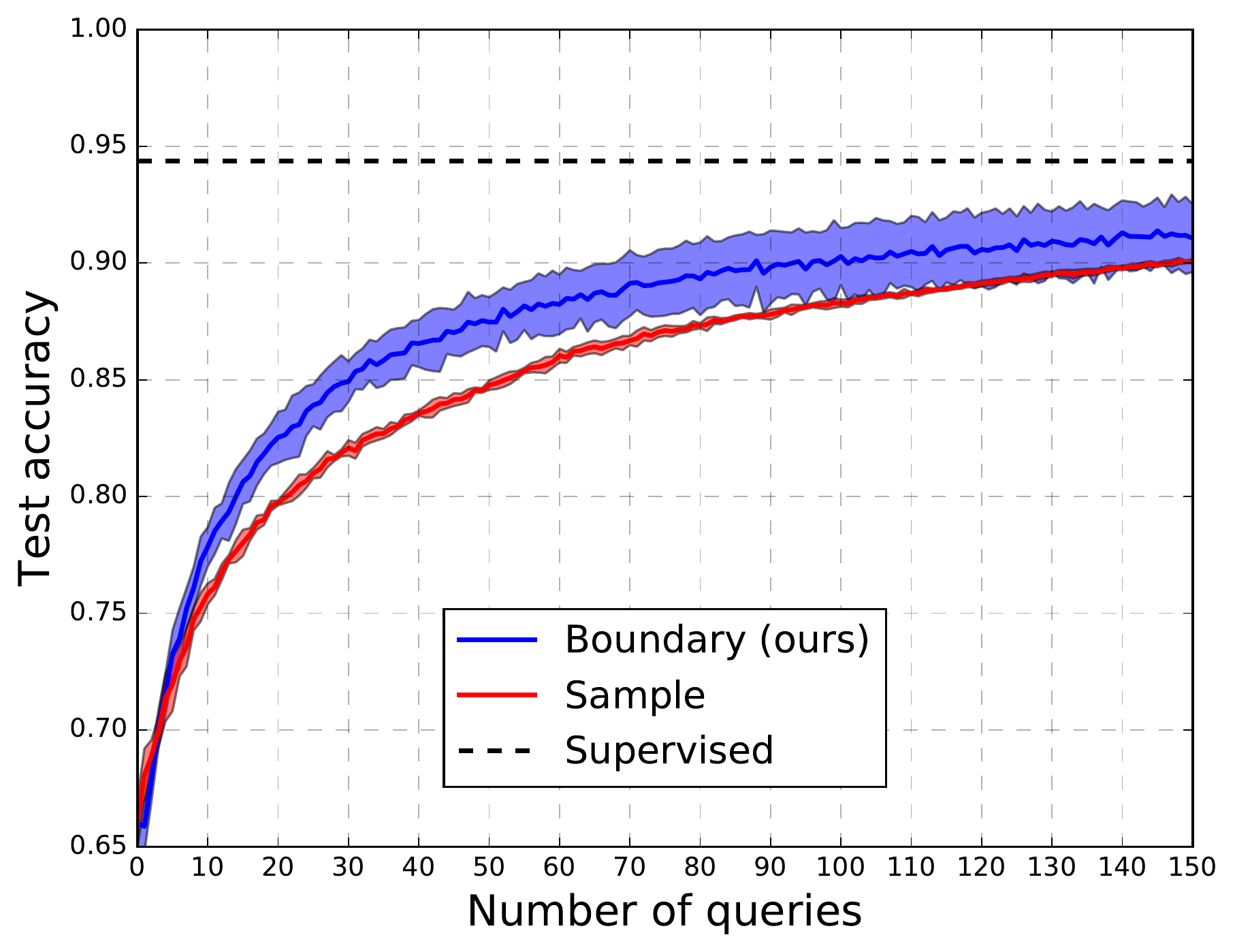}
   \caption{SVHN averaged over all classes.}
   \label{}
\end{subfigure}
\begin{subfigure}{0.33\textwidth}
   \includegraphics[width=1\linewidth]{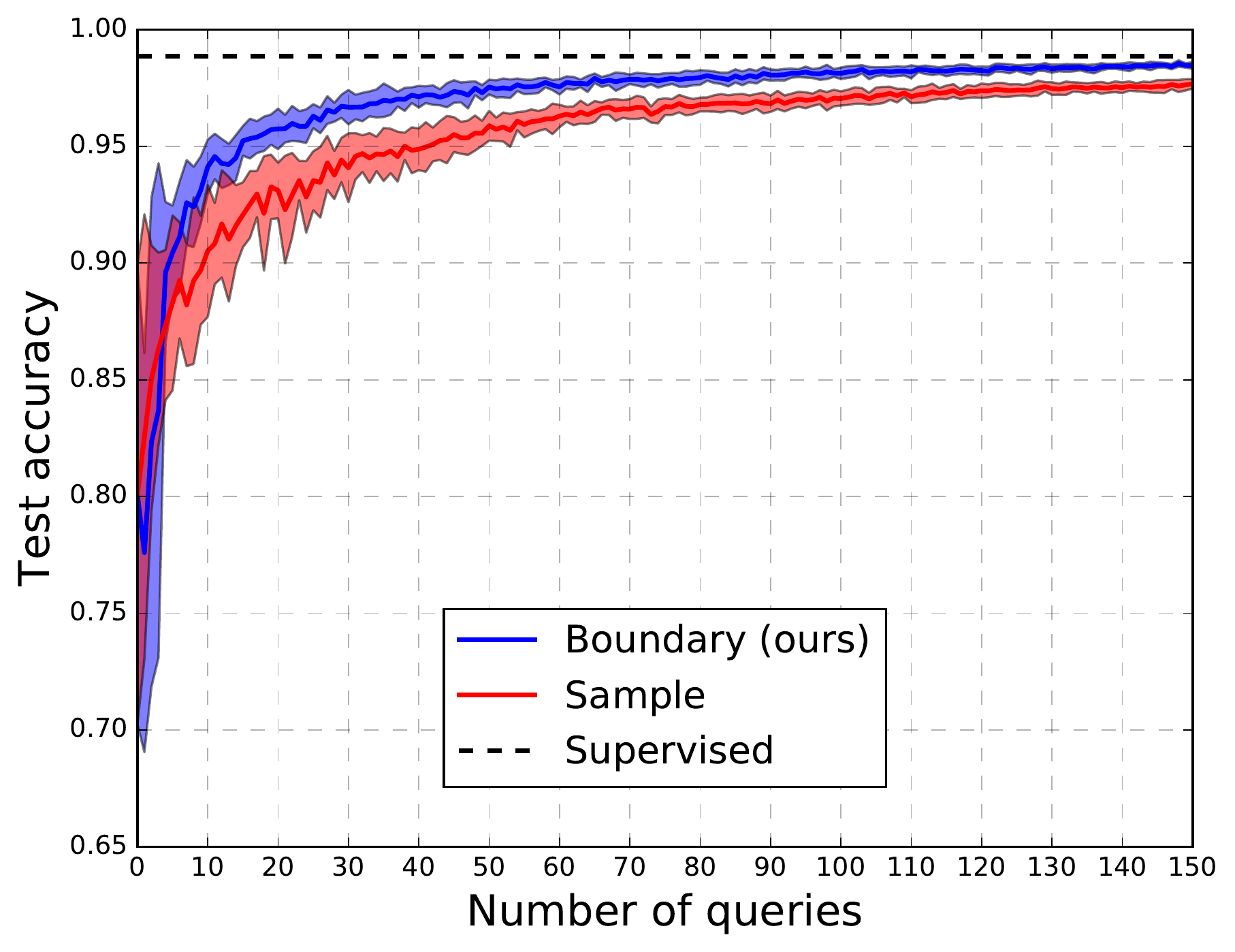}
   \caption{Shoe-Bag both classes.}
   \label{}
\end{subfigure}
\caption{Learning curves over all datasets and all class pairs using uncertainty sampling as query strategy. The experiments are repeated 5 times, standard deviations are indicated by line width. The fully supervised oracle-SVM is the upper bound. Our boundary method outperforms the sample-based method.}
\label{fig:generalizationClasses}
\end{figure*}

The results in~\tab{humanOracle} show that an oracle-SVM outperforms a human annotator. This is probably because the active learner method that is being trained is also an SVM, and since the oracle is also an SVM it will choose the perfect samples. For humans, boundary annotation always improves over sample annotation. Yet, for SVHN and Shoe-Bag this improvement is not significant. This is probably due to the small number of queries, where our method after only 10 iterations has not yet achieved peak performance as corroborated by the learning curves in~\fig{generalizationClasses}. 

\subsection{Exp 5: Generalization over classes}
Up to now we have shown that our method outperforms sample-based active learning on a subset of MNIST and SVHN. To see whether our method generalizes to the other classes we evaluate the performance averaged over all the SVHN and MNIST class pairs using uncertainty sampling as query strategy. We show  results in~\tab{generalizationClasses} and plot the learning curves in~\fig{generalizationClasses}. 
Averaged over all datasets and class pairs our method is significantly better than the sample-based approach.

\begin{figure}
\centering
\includegraphics[width=0.8\linewidth]{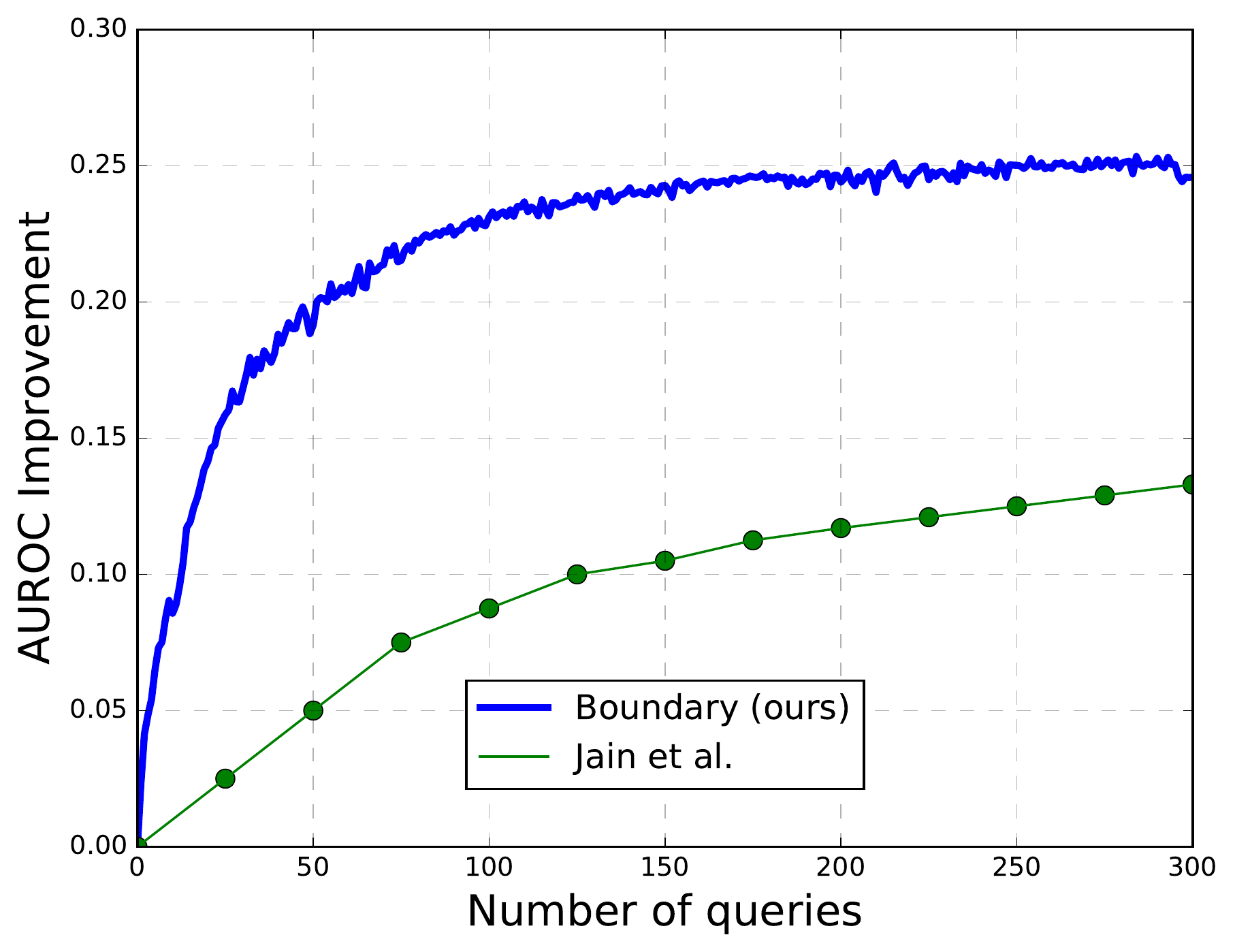}
\caption{Comparison of our method against Jain \etal~\cite{jain2010hashing} on CIFAR-10 using 384-d GIST descriptors. The results are averaged over 5 repeats. We outperform Jain \etal~\cite{jain2010hashing}}
\label{fig:jain_plot}
\end{figure}

We also evaluate on the CIFAR-10 dataset using GIST features following Jain \etal (EH-Hash)~\cite{jain2010hashing} for uncertainty sampling. Results in~\fig{jain_plot} show we clearly outperform EH-Hash in terms of AUROC improvement.

\begin{table}
\begin{tabular}{l l l} 
\multicolumn{3}{c}{Experiment 5: Full dataset evaluation} \\ \toprule
& Sample & Boundary (ours) \\ \cmidrule(lr){2-2} \cmidrule(lr){3-3}
MNIST & $147.8 \pm 0.06$& \textbf{148.3 $\pm$ 0.04}\\
SVHN & $127.8 \pm 0.2$& \textbf{130.9 $\pm$ 1.9} \\
Shoe-Bag & $143.2 \pm 0.6$ & \textbf{145.4 $\pm$ 0.5} \\
\bottomrule
\end{tabular}
\caption{AULC results for sample-based active learning and boundary active learning for all datasets after 150 queries (maximum possible score is 150), averaged over all class pairs. The experiments are repeated 5 times and significant results are shown in bold. Significance is measured with a paired t-test with p $<$ 0.05. For all datasets our method significantly improves over sample-based active learning. }
\label{tab:generalizationClasses}
\end{table}

%% file: conclusion.tex
\section{Discussion}

We extend active learning with a method for direct decision boundary annotation. We use a deep generative model to synthesize new images along a 1-dimensional query line, and ask an oracle to annotate the point where the images change class: this point is an annotation of the decision boundary. Note that this may not lead to a direct acceleration of the annotation speed. If the annotation cost is per hour, \eg a radiologist, then ease of annotation and system speed become key. If costs are per task, e.g. Amazon Mechanical Turk, then the informativeness of each annotation should be maximally exploited. Our method falls in the latter category: We increase the informativeness per annotation. 

One disadvantage of our method is that it is very easy to annotate changes visually, but this is not so straightforward in other domains. The core of our method can in principle also be used on any input data, but actually using a human oracle to detect a class change for non-visual data would become tedious fast. For example, having a human annotate a class-change for raw sensor data, speech or text documents would be quite difficult in practice. Our method could still be applicable if the non-visual data can be easily visualized. 

Another problem is precisely annotating the decision boundary when the margin between classes is large. With a large margin the generated samples may all look similar to each other and it is difficult to annotate the class change exactly. A solution could be to annotate the margin on each side instead of the decision boundary. 

Our method depends critically on the quality of the generative model. We specifically evaluated this by including the Shoe-Bag dataset where the quality of the generated samples impairs the consistency of human annotation. If the generative model is of low quality, our method will fail as well. Deep generative models are an active area of research, so we are confident that the quality of generative models will improve. One possible direction for future work could be to exploit the knowledge of the annotated decision boundary to update the generative model. 

In this work we consider linear models only. The decision boundary hyperplane lives in a K-1 dimensional space, and thus K-1 independents points span the plane exactly. The reason why we need more than K-1 query annotations is that boundary annotation points may not be independent. Future work on a new query strategy that would enforce boundary point independence may be promising to reduce the number of annotations required. For non-linear models, a non-linear 1-dimensional query line could perhaps work better. Also, when data sets are not linearly separable we may require more than one annotation of the decision boundary for 1 query line. This is left for future work.

Our paper shows that boundary annotation for visual data is possible and improves results over only labeling query samples.  We show that our method can plug in existing active learning strategies, that humans can consistently annotate the boundary if the generative model is good enough, that our method is robust to noise, and that it significantly outperforms sample-based methods for all evaluated classes and data sets.

%% file: 1673-supp.tex
\appendix

\section{Project query point on decision boundary}
\label{app:appendix_projection}

\begin{figure}[h!]
\centering
\includegraphics[width=0.5\linewidth]{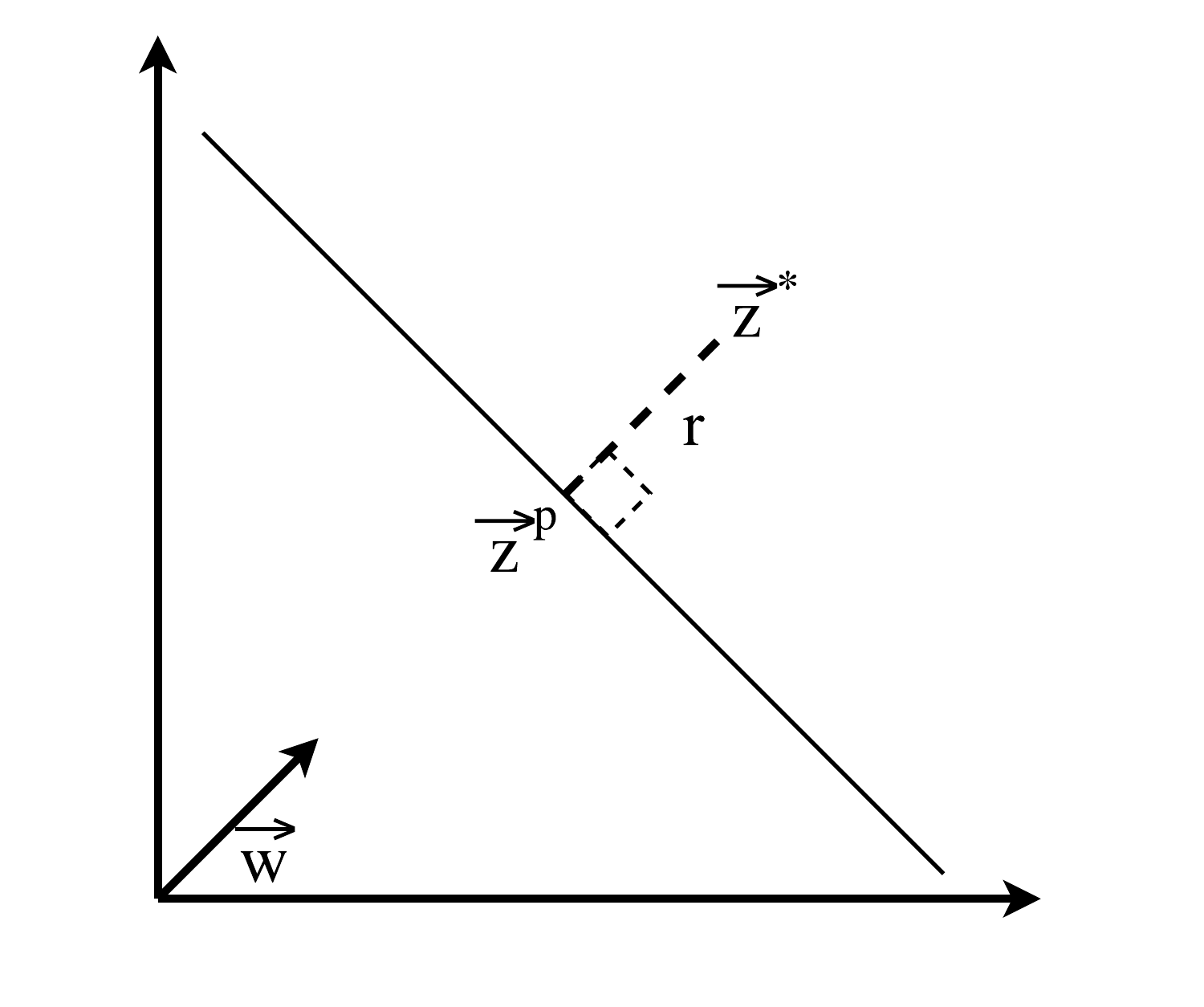}
\caption{$\mathbf{z}^p$ is the projection of query point $\mathbf{z}^*$ on the decision boundary.}
\label{fig:projecting}
\end{figure}

Projecting query point $\mathbf{z}^*$ on the decision boundary parameterized by $\mathbf{w}$ and $b$ yields $\mathbf{z}^p$, see~\fig{projecting}. 

\noindent The projected point is defined as
\begin{equation}
\mathbf{z}^p = \mathbf{z}^* - y r \frac{\mathbf{w}}{|\mathbf{w}|},
\label{eq:proj1}
\end{equation}
where $y$ is the class label of $\mathbf{z}^*$. Furthermore, $\mathbf{z}^p$ lies on the decision boundary and thus satisfies 
\begin{equation}
\mathbf{w}^\intercal\mathbf{z}^p + b = 0. 
\label{eq:proj2}
\end{equation}
Substituting~\Eq{proj1} in~\Eq{proj2} yields:
\begin{equation}
\mathbf{w}^\intercal\left(\mathbf{z}^* -  y r \frac{\mathbf{w}}{|\mathbf{w}|}\right) + b = 0.
\end{equation}
Now, we solve for $r$:
\begin{equation}
r = \frac{\mathbf{w}^\intercal \mathbf{z}^* + b}{y |\mathbf{w}|}.
\label{eq:proj3}
\end{equation}
Substituting $r$ from~\Eq{proj3} in~\Eq{proj1} gives:
\begin{equation}
\begin{split}
\mathbf{z}^p &= \mathbf{z}^* - y \frac{\mathbf{w}^\intercal \mathbf{z}^* + b}{ y |\mathbf{w}|} \cdot \frac{\mathbf{w}}{|\mathbf{w}|}\\
&= \mathbf{z}^* - \frac{(\mathbf{w}^\intercal \mathbf{z}^* + b)\mathbf{w}}{ |\mathbf{w}| |\mathbf{w}|}\\
&= \mathbf{z}^* - \frac{(\mathbf{w}^\intercal \mathbf{z}^* + b)\mathbf{w}}{\mathbf{w}^\intercal\mathbf{w}}.
\end{split}
\end{equation}

\section{Results measured in Average Precision}
\label{app:resultsInAP}

As requested by one of our anonymous reviewer we also evaluate our results measured in Average Precision, see Tables~\ref{tab:queryStrategies_ap}, \ref{tab:labelnoise_ap}, \ref{tab:humanOracle_ap}, \ref{tab:generalizationClasses_ap}.

\begin{table*}
\begin{tabular}{l ll ll ll}
 \multicolumn{7}{c}{ Experiment 1: Evaluating various query strategies (average precision)} \\ \toprule
 & \multicolumn{2}{c}{MNIST 0 vs. 8} & \multicolumn{2}{c}{SVHN 0 vs. 8} & \multicolumn{2}{c}{Shoe-Bag} \\
Strategy & Sample & Boundary (ours) &   Sample & Boundary (ours) &   Sample & Boundary (ours)  \\ \cmidrule(lr){1-1} \cmidrule(lr){2-3} \cmidrule(lr){4-5} \cmidrule(lr){6-7}
 Uncertainty  & $98.9 \pm 0.2$ & \textbf{99.4 $\pm$ 0.2} & $91.1 \pm 0.8$ & \textbf{93.8 $\pm$ 0.7} & $98.9 \pm 0.2$ & \textbf{99.3 $\pm$ 0.1}\\
 Uncertainty-dense & $94.5 \pm 11.0$ & \textbf{96.5 $\pm$ 10.7} & $77.1 \pm 5.7$ & \textbf{89.4 $\pm$ 2.1} & $84.6 \pm 4.0$ & \textbf{96.9 $\pm$ 1.1} \\
5 Cluster centroid & $98.6 \pm 0.1$ & \textbf{99.6 $\pm$ 0.03} & $75.5 \pm 5.4$ & \textbf{83.1 $\pm$ 1.2} & $95.1 \pm 0.8$ & \textbf{99.3 $\pm$ 0.1} \\
Random & $98.4 \pm 0.4$ & \textbf{99.2 $\pm$ 0.3} & $89.3 \pm 1.4$ & \textbf{93.7 $\pm$ 0.8} & $98.1 \pm 0.4$ & \textbf{99.3 $\pm$ 0.1}\\
\bottomrule
\end{tabular}
\caption{Average precision averaged over 150 iterations. }
\label{tab:queryStrategies_ap}
\end{table*}

\begin{table*}
\centering
\begin{tabular}{l ll ll ll} 
\multicolumn{7}{c}{Experiment 3: Evaluating annotation noise} \\ \toprule
 Sampling noise& \multicolumn{2}{c}{MNIST 0 vs. 8} & \multicolumn{2}{c}{SVHN 0 vs. 8} & \multicolumn{2}{c}{Shoe-Bag} \\
(\# images) & Sample & Boundary (ours) &   Sample & Boundary (ours) &   Sample & Boundary (ours)  \\ \cmidrule(lr){1-1} \cmidrule(lr){2-3} \cmidrule(lr){4-5} \cmidrule(lr){6-7}
0 & $99.0 \pm 0.2$ & \textbf{99.5 $\pm$ 0.1} & $90.9 \pm 1.4$ & \textbf{93.7 $\pm$ 0.8} &  $98.9 \pm 0.2$ & \textbf{99.3 $\pm$ 0.2}\\
1 & $99.0 \pm 0.2$ & \textbf{99.5 $\pm$ 0.1} & $90.9 \pm 1.4$ & \textbf{93.4 $\pm$ 0.7} & $98.9 \pm 0.2$ & \textbf{99.3 $\pm$ 0.2}\\
2 & $99.0 \pm 0.2$ & \textbf{99.4 $\pm$ 0.2} & $90.9 \pm 1.4$ & \textbf{92.3 $\pm$ 1.6} & $98.9 \pm 0.2$ & $99.1 \pm 0.6$\\
3 & $99.0 \pm 0.2$ & \textbf{99.3 $\pm$ 0.1} & $90.9 \pm 1.4$ & \textbf{92.5 $\pm$ 0.8} & $98.9 \pm 0.2$ & $99.1 \pm 0.6$\\
4 & $99.0 \pm 0.2$ & \textbf{99.1 $\pm$ 0.1} & $90.9 \pm 1.4$ & $91.3 \pm 0.9$ & $98.9 \pm 0.2$ & \textbf{99.1 $\pm$ 0.3}\\
5 & $99.0 \pm 0.2$ & $99.0 \pm 0.1$ & $90.9 \pm 1.4$ & $86.7 \pm 9.7$ & $98.9 \pm 0.2$ & $98.8 \pm 0.4$\\
\bottomrule
\end{tabular}
\caption{Average Precision results for noisy boundary active learning with uncertainty sampling for MNIST (classifying 0 and 8), SVHN (classifying 0 and 8) and Handbags vs. Shoes averaged over 150 queries (maximum possible score is 150). Each experiment is repeated 15 times. For each row, the significantly best result is shown in bold, where significance is measured with a paired t-test with p $<$ 0.05. Noise has been added to the boundary annotation points; not to the image labels.} 
\label{tab:labelnoise_ap}
\end{table*}

\begin{table*}
\begin{tabular}{l ll ll ll} 
\multicolumn{7}{c}{Experiment 4: Evaluating a human oracle} \\ \toprule
  & \multicolumn{2}{c}{MNIST 0 vs. 8} & \multicolumn{2}{c}{SVHN 0 vs. 8} & \multicolumn{2}{c}{Shoe-Bag} \\
Annotation & Sample & Boundary (ours) &   Sample & Boundary (ours) &   Sample & Boundary (ours)  \\ \cmidrule(lr){1-1} \cmidrule(lr){2-3} \cmidrule(lr){4-5} \cmidrule(lr){6-7}
Human oracle & - & - & $63.0 \pm 7.2$ & $64.6 \pm 7.6$ & - & - \\
SVM oracle & $94.7 \pm 2.5$ & \textbf{96.1 $\pm$ 2.7} & $74.0 \pm 5.8$ & $74.9 \pm 5.4$ & $94.9 \pm 1.5$ & \textbf{95.8 $\pm$ 1.6}\\
\bottomrule
\end{tabular}
\caption{Average Precision results for a human and a SVM oracle for sample-based active learning and our boundary active learning for MNIST (classifying 0 and 8), SVHN (classifying 0 and 8) and Shoe-Bag averaged over 10 queries (maximum possible score is 10). The experiments are repeated 15 times and  significant results per row are shown in bold for p $<$ 0.05.} 
\label{tab:humanOracle_ap}
\end{table*}

\begin{table}[h]
\centering
\begin{tabular}{l l l} 
\multicolumn{3}{c}{Experiment 5: Full dataset evaluation} \\ \toprule
& Sample & Boundary (ours) \\ \cmidrule(lr){2-2} \cmidrule(lr){3-3}
MNIST & $99.78 \pm 0.02$& \textbf{99.84 $\pm$ 0.03}\\
SVHN & $90.65 \pm 0.19$& $91.22 \pm 1.95$ \\
Shoe-Bag & $98.89 \pm 0.18$ & \textbf{99.34 $\pm$ 0.13} \\
\bottomrule
\end{tabular}
\caption{Average Precision results for sample-based active learning and boundary active learning for all datasets averaged over 150 queries (maximum possible score is 150), averaged over all class pairs. The experiments are repeated 5 times and significant results are shown in bold. Significance is measured with a paired t-test with p $<$ 0.05. }
\label{tab:generalizationClasses_ap}
\end{table}